\documentclass[11pt]{article}
\usepackage{coling2020,times,hyperref,natbib,booktabs,microtype,xcolor,graphicx,mdwlist}

\colingfinalcopy

\definecolor{darkblue}{rgb}{0, 0, 0.5}  
\hypersetup{colorlinks=true,citecolor=darkblue, linkcolor=darkblue, urlcolor=darkblue}

\title{Results of a Single Blind Literary Taste Test \\
    with Short Anonymized Novel Fragments}  

\author{Andreas van Cranenburgh \\
  CLCG, University of Groningen \\
  {\tt a.w.van.cranenburgh@rug.nl} \\ \And
  Corina Koolen \\
  Independent Scholar \\
  {\tt corinakoolen@gmail.com}}

\date{}

\begin{document}
\maketitle
\begin{abstract}
    It is an open question to what extent perceptions of literary quality
    are derived from text-intrinsic versus social factors. While supervised
    models can predict literary quality ratings from textual factors quite
    successfully, as shown in the Riddle of Literary Quality
    project \citep{koolen2020survey}, this does not prove that social factors
    are not important, nor can we assume that readers make judgments on
    literary quality in the same way and based on the same information
    as machine learning models.    
    We report the results of a pilot study to gauge the
    effect of textual features on literary ratings of Dutch-language novels by
    participants in a controlled experiment with 48 participants. In an
    exploratory analysis, we compare the ratings to those from
    the large reader survey of the Riddle in which social factors were not
    excluded, and to machine learning predictions of those literary ratings.
    We find moderate to strong correlations of questionnaire ratings with the
    survey ratings, but the predictions are closer to the survey ratings.
    Code and data: \url{https://github.com/andreasvc/litquest}
\end{abstract}

\section{Introduction}
It remains an open question why some novels are considered literary,
while other novels such as thrillers are considered
less prestigious. One position in this debate is that formal
features of the text play an important role, i.e., formalism.
The notion that foregrounding language is involved with the
aesthetic appreciation of literary texts is supported
by experimental studies~\citep[e.g., ][]{hakemulder2004foregrounding}.
Recent work investigates the role of the text in literary quality empirically
on a larger scale by collecting literary ratings of novels in a large reader
survey~\citep[henceforth the Riddle survey, ][]{koolen2020survey}. This survey
provides a way to investigate the association
of literary judgments and textual features directly.
Can human judgments of literary quality be predicted by machine learning models
from purely textual features? The answer turns out to be yes, to a substantial
extent. The model by \citet{vancranenburgh2017literary} showed that textual
features can explain 61\% of the variation in the literary ratings.

However, machine learning may pick up on various subtle
frequency differences that humans do not notice; compare the
case of authorship attribution, which is arguably more difficult
to do by humans by hand, than it is for a computer with a simple
table of function word frequencies and cosine distances.
In order to place the success of the machine learning model
that can predict literariness in context, we would have to ask
humans to assign ratings purely based on a text fragment,
without being influenced by the prestige associated with an author's name or
the title of the novel.
The setup of such a `challenge' was proposed in
\citet{vancranenburgh2019pepsi}; however, no results were presented.
Wine experts perform at chance level in blind taste
tests~\citep{hodgson2008wine}; will readers be able to recognize literature?

In this paper we report on the results of a pilot
version of such an experiment and analyze the results.
This allows us to see the extent to which our participants
agree with ratings in the Riddle survey who rate based on author and title,
as well as how they compare to the machine learning models that perform a
similar task. This also allows us to estimate
the importance of textual features for human readers
when making judgments.
By looking into the motivations that the participants have
given for their ratings, we also hope to identify interesting
linguistic features of literary quality, which can be investigated
further in future computational and literary studies work.

\section{Questionnaire setup}
We selected 8 originally Dutch novels for the pilot; 4 of which were rated as
highly literary (5.5--7.0) in the Riddle survey, and 4 rated as neither
non-literary or very literary (3-5--5.0) in that survey.
We deliberately did not select non-literary novels with low ratings of less
than 3.5, because previous research showed that the genre of such novels is
very easy to recognize \citep{vancranenburgh2017literary},
and we are interested in the more subtle differences between literary
novels and `not quite literary' novels.
In addition, author gender is balanced, both for the literary and
the not quite literary novels.

From these novels we selected fragments of 250 words,
from the beginning of a chapter, though not the first chapter.
We rule out the influence of author prestige by not showing any
metadata of the fragments that are presented to the participants.
In addition to not presenting metadata, we also anonymize the fragments;
i.e., names of the main characters in the text are abbreviated to initials.
Fragments are presented in an arbitrary order (although we
did not shuffle the fragments across participants).

We use the same 7-point Likert scale as in the Riddle survey,
which ranges from not at all literary to very literary.
As in that survey, we did not give a definition of what `literary' is,
because we did not want to influence the participants.
For a given fragment, participants were asked the following
questions:

\begin{enumerate*}
    \item ``How literary do you think this fragment is?'' (7-point Likert scale)
    \item ``Briefly explain why you chose this rating'' (text field)
    \item ``If a specific phrase contributed to this judgment, cite it'' (text field)
\end{enumerate*}

We recruited participants 
from our social network.
They range from casual to enthusiastic readers, of various genres,
with ages ranging from 30--60.
We collected responses from 48 participants.
Compared to the 14k participants of the Riddle survey,
this is a small pilot.
However, for the Riddle survey, we look only at the ratings of books the
participants had rated, which means that the number of ratings varies per book.
It was shown that 50--100 ratings are sufficient to compute a reliable
mean~\citep{vancranenburgh2017literary}.
In contrast, all fragments were rated by
all 48 participants in our questionnaire.

\begin{table}[t]\centering
\begin{tabular}{llllrrr}\toprule
     &                         &          & Author  & \multicolumn{2}{c}{Rating} & Riddle \\
Rank & Novel                   & Genre    & gender & M   & SD & rank \\ \midrule
1 & Mortier, Godenslaap        & literary & male   & 5.8 & 1.2 & 2 \\
2 & Durlacher, De Held         & literary & female & 4.9 & 1.4 & 56 \\
3 & Dijkzeul, Gouden Bergen    & suspense & female & 4.7 & 1.3 & 156 \\
4 & Den Tex, Wachtwoord        & suspense & male   & 4.6 & 1.4 & 147 \\
5 & Smit, Vloed                & literary & female & 4.3 & 1.6 & 103 \\
6 & Van der Heijden, Tonio     & literary & male   & 4.3 & 1.3 & 7 \\
7 & Dorrestein, De stiefmoeder & literary & female & 4.0 & 1.7 & 64 \\
8 & Appel, Van twee kanten     & suspense & male   & 3.4 & 1.5 & 184 \\ \bottomrule
\end{tabular}
\caption{Results of the questionnaire with a comparison
    to the ranking in the Riddle survey (401 novels).}\label{tblresults}
\end{table}

\section{Literary ratings}
In this first analysis, we consider the mean literary rating for each novel
fragment in the questionnaire. We compare these means to the ratings in the
Riddle survey and to predicted ratings from supervised models.

\subsection{Comparison with the large reader survey}
\autoref{tblresults} presents the results as a ranking, ordered by the rating
in the questionnaire. We note 4 differences in ranking. 
Compared to the Riddle survey, the thrillers at rank 3 and 4 are rated as more
literary, while the literary novels at rank 6 and 7 are rated substantially
less literary by the participants. The gap between male and female authors is
less pronounced than in the Riddle survey, where literary female authors
were rated 0.5 points lower than male authors, on
average~\citep{koolen2018reading}. Similarly, the genre effect of suspense
novels being rated lower than literary novels disappears
for the novels by Dijkzeul and Den Tex.
That their writing stands out is not surprising; Dijkzeul has been nominated
several times for the \emph{Gouden Strop}, the most well-known award for Dutch
suspense novels, while Den Tex is a three time laureate of the award.

It is possible that these differences are due to author prestige 
not playing a role in the questionnaire, but it may also be due to the
difference in rating a small fragment on stylistic aspects versus a complete
novel where plot may also play a role.
Rating a small fragment in isolation is very different from rating a a novel
one has read from memory, based only on the author and title. 

When a reader rates a novel from memory, social factors will most likely come
into play. If an author is known to have been acknowledged by literary
institutions, for example through literary prizes and literary studies, this
will add to their prestige~\citep{verboord2003prestige}. And in turn, this
prestige could influence a reader to award a higher rating on a scale of
literary quality. Alternatively, a lack of prestige might result in a lower
judgment, as is the case with suspense novels. Respondents in the National
Reader Survey would explain a low rating for a suspense novel with comments
such as ``It's a suspense novel,'' signifying a tendency to associate lower
literary quality with a specific genre.
On the other hand, novels by female authors are seen as less literary overall,
according to our larger reader survey, even when the novel judged won a
literary prize~\citep{koolen2020survey}. In sum, prestige---of an author and/or
novel---constitutes a complex web of factors, which likely plays a part in
literariness judgments, as the results of the questionnaire show.
Female authors and suspense authors are no longer separated in literary
judgment from the male and literary authors, respectively.

There are problems with this assumption, however. For example, the literary
novel by female author Durlacher has a suspenseful plot which might have
resulted in lower literary ratings in the Riddle survey. At the same time the
writing style was praised as layered and literary, which may explain the high
rating in the questionnaire---since the latter mainly invites judgments based
on writing style. Dijkzeul's suspense novel is another example. One reviewer
calls the plot of this novel
predictable;\footnote{\url{https://www.hebban.nl/recensie/a-vd-heide-over-gouden-bergen}}
this could explain why it is rated lower in the Riddle survey than in our
questionnaire, where plot plays no role. Further research is necessary to
confirm these explanations.

However, it could also be that the scales are not comparable.
Overall it seems that the participants in our questionnaire are more
conservative, because only one novel got a rating above 5.
This may be due to difficulty of rating such short fragments,
which might be compared to judging a wine by tasting a drop of it;
in the absence of sufficient information, we observe a regression to the mean.

\subsection{Comparison with machine learning models}
\begin{figure}[t]\centering
    \includegraphics[width=0.59\linewidth]{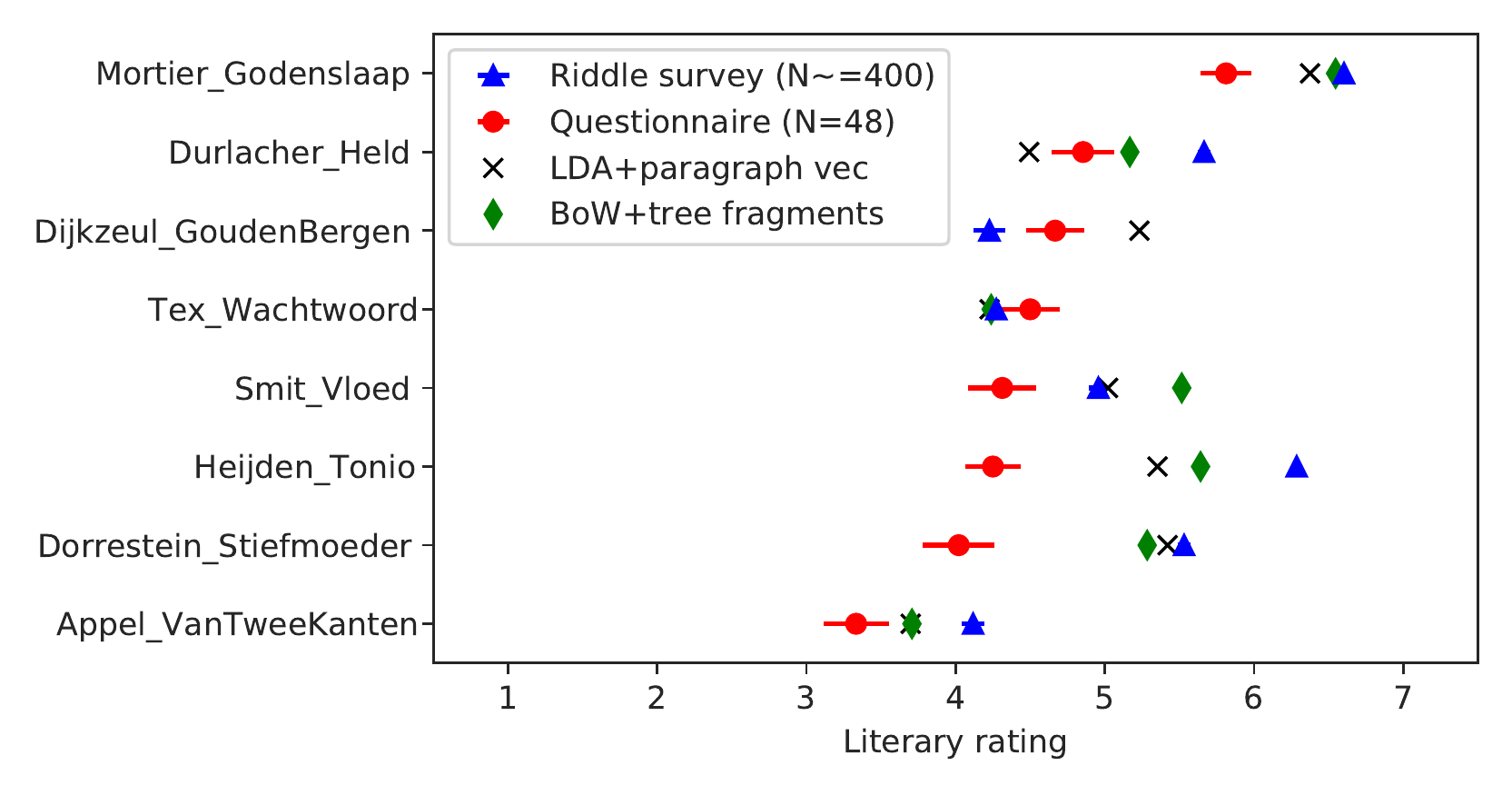}~%
\includegraphics[width=0.4\linewidth]{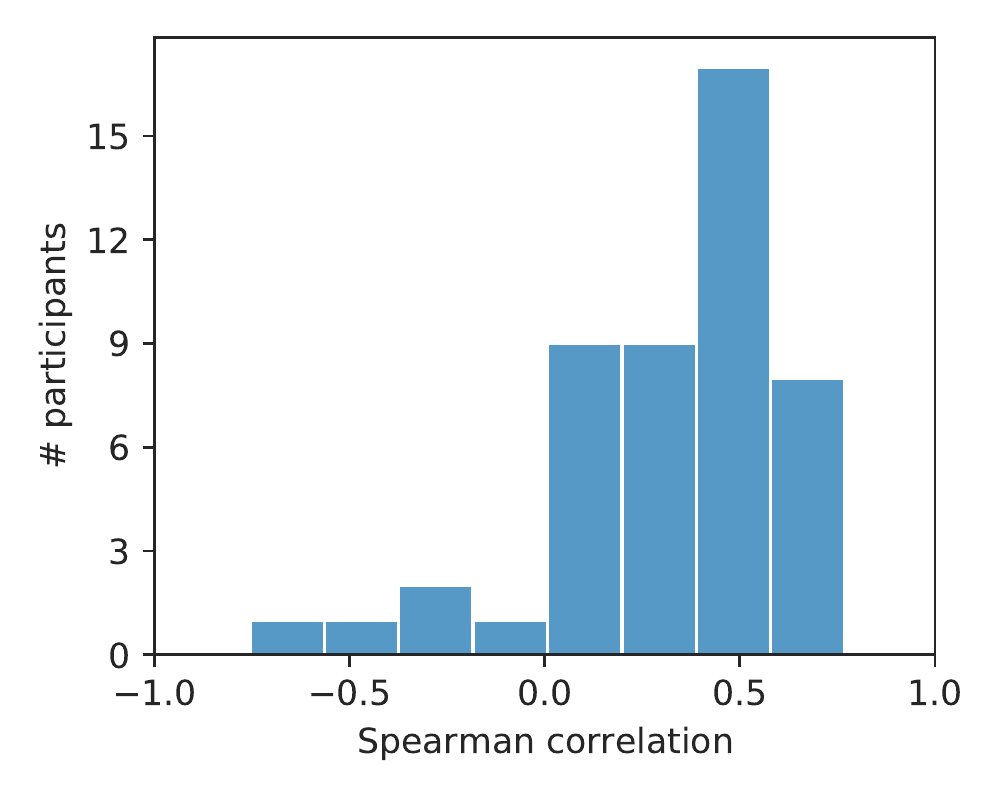}
    \caption{%
    Left:
    Comparison of ratings from the questionnaire, Riddle survey and predictions from machine learning models.
    For the Riddle survey the average number of ratings is indicated.
    Error bars on ratings show standard error (note that
    the error bars of the Riddle survey are so small they are barely visible).
    Right:
    A histogram of Spearman correlations, showing
    how strongly the responses of the participants in the questionnaire are correlated with the Riddle survey.
    }\label{figresults}
\end{figure}

The plot on the left of \autoref{figresults} shows a comparison
of the ratings in our questionnaire, the Riddle survey,
and two predictive models from previous work.
The predictive models take the mean ratings from the Riddle survey as
ground truth by training and evaluatinng on them using crossvalidation.

The first model, ``BoW + tree fragments,''
uses a large number of textual features including word bigrams and
syntactic tree fragments
and is based on long novel fragments of 1000 sentences~\citep{vancranenburgh2017literary};
this model reached an $R^2$ of 61.1.
This score was obtained using crossvalidation of a regularized linear SVM
regression model. An $R^2$ score is a measure of the variation explained and
ranges from 0 to 100, with 0 being baseline performance and 100 being a perfect
score. Tree fragments where extracted based on their frequency and correlation
with literary ratings in the training fold.

The second model, ``LDA + paragraph vectors,'' addresses the harder task
of predicting literary ratings from short fragments of 1000 words
and only uses LDA topic weights~\citep{blei2003latent} and DBOW paragraph
vectors~\citep{le2014doc2vec} based on the 401 novels as document
representations~\citep{vancranenburgh2019vecspace};
this model reached an $R^2$ of 52.2.
This score was also obtained using crossvalidation, with a regularized linear
model (Ridge regression). The topic weights are obtained using Mallet, while
the paragraph vectors were obtained with Gensim.

For the latter model, we have identified the matching 1000 word
fragment containing the 250 word fragment presented in the
questionnaire, and show the corresponding prediction in \autoref{figresults}.
While the comparison would be even more faithful if we evaluated
the prediction on the exact same 250 word fragments, we opt for using
the existing tuned and benchmarked model.
If we take the Riddle survey as ground truth,
the predictions of the models are closer than the ratings in our questionnaire,
with two exceptions, Durlacher and Dijkzeul.
This suggests that
the predictive models may pick up on cues
in the text that are not apparent to an untrained observer.
However, it could also be that the participants in the questionnaire
are more accurate because they were not influenced by social factors.

We can also compare our questionnaire and the Riddle survey in terms of
the participants. How similar are their ratings?
Because we are not interested in the precise differences in
ratings, we want to focus on their rankings. We therefore compute
a Spearman correlation coefficient of the ratings of each participant
in the questionnaire with the mean ratings in the Riddle survey for
the 8 novels.
However, due this limited number of 8 data points per participant,
these correlations are only presented as an exploratory analysis
to give an indication of effect sizes, 
and we do not perform a formal hypothesis test with p-values.
The plot on the right of \autoref{figresults}
shows a histogram of these correlations, showing
that most participants of the questionnaire displayed a moderate to
large agreement ($\rho>0.4$) with the Riddle survey.
These correlations support the construct validity of literariness as a
measurable variable, since the human ratings from very different
experiments agree to a large extent. If the participants would have
`failed the challenge,' we would expect the distribution of
correlations to be centered around 0, but the majority shows a
positive correlation.
In the Riddle survey, participants had read the complete novel they rated,
but may also have been influenced by
author prestige. 
In our questionnaire, participants were only presented with a
short fragment, but were not influenced by other factors (except in the case
that the participant recognized an author or novel from the fragment, which is
not highly likely considering the briefness of the fragments and the
anonymization of the names of the main characters).

\section{Analysis of motivations}
Overall, the common thread in the motivations is that elaborate
and sophisticated language is seen as more literary. Most motivations
refer to specific stylistic aspects. Participants also remarked that
the task was hard due to the short text fragments.
The following sentence was cited by multiple participants who
agreed on its high literariness (our translation):
``She brings out the world's blissful mongoloid smile, the grinning, wet
shining zen of dumb objects, from which she blows the names off like chaff.''

However, the choice of fragment likely plays a large role.
The fragments were selected arbitrarily, and are not necessarily
representative of the whole novel. Especially for Dorrestein,
this likely played a role. In the fragment, a woman observes a group of
children, followed by the following free indirect discourse  (our translation):
``She notices that she moves her jaws.
I am Mrs Pacman. Bite, swallow, gone. Bite swallow gone.''
This phrase struck many participants, but it is striking that
they disagreed on whether it made the fragment more or less
literary. One participant cited this phrase as motivation for a
6 out of 7 (`no fixed pattern, surprising, would like to read more'),
while another cited it to give a rating
of 1 out of 7  due to the sudden shift in style
(`a hodgepodge of old fashioned words and modern imagery').
Incidentally, such examples may be an issue for the notion of foregrounding
as an explanation of literariness:
while we find phrases that stand out for many of the participants,
they do not agree on their effect on the literary status of the
fragment. This requires further study; for example, would the participants'
judgment change if more context is given?

While the fragment by Den Tex scored high in the questionnaire,
several of the participants cited the mention of a soccer player,
or the fact that soccer was discussed in the fragment,
as a reason for judging the fragment as non-literary.
This suggests that there is a perception that references to low brow, popular
culture are associated with lower prestige, while literature should be about
more timeless and serious matters.
We can hypothesize that this is part of a heuristic:
by default, references to low-brow popular culture trigger
a low rating; on the other hand,
such references can also appear in high literature, but in this case
more context (i.e., a longer fragment)
may be needed to see how such references fit in the story. 

\section{Discussion and Conclusion}
We found a reasonable consensus for the strongest style differences
in the fragments.
The correlations of the Riddle survey ratings with our questionnaire ratings
range from moderate to strong, which supports the construct validity
of literariness as a variable.
The difference between the ranking of the
most and least literary fragment was preserved, and some of the
changes in ranking could be explained as genre and gender effects
that disappear when rating an anonymized fragment.

Participants found the task of rating short fragments hard,
and the predictions by the machine learning models
are closer to the survey ratings.
However, the machine learning models are trained,
while the participants are not.
The ability of predictive models to reproduce
the quality ratings should be interpreted carefully
since the models may pick up on more than just literariness
from the textual features, such as the aforementioned
genre and gender effects.

The participants agree on criteria (e.g., word usage),
agree on salience of phrases, but sometimes disagree on how literary they are.
This suggests that foregrounded language is not necessarily seen as literary
language, and raises the question of how to identify potential linguistic
markers of literary language for further study.

Our work is related to work on experimental aesthetics. In one branch of
experimental aesthetics based on the ideas of \citet{fechner1876vorschule}, the
subjects' ideas on aesthetics are researched. Respondents are asked to recall
as many adjectives related to aesthetic judgment as possible, for instance of
literary novels. In \citet{knoop2016mapping}, this approach leads to the
conclusion that `beautiful' and `suspenseful' are central to descriptions of
aesthetic judgment of fictional literature. This type of research is related to
the influence of the text, but \citet{knoop2016mapping} did not ask readers to
reflect on specific works. Other work on experimental aesthetics considers
foregrounding effects and does present participants with novel
fragments~\citep[e.g., ][]{hakemulder2004foregrounding}
or lines of poetry \citep[e.g., ][]{blohm2018sentence}. 
However, more generally, there is more to literary quality than aesthetics.
Even though there has been a focus on the aesthetic experience of literary
language \citep[cf.~][]{peer2008quality}, ethical, moral, affective and other
motivations can play a role in literary judgments as well.

In future work, we want to determine the influence of
author prestige by conducting a controlled experiment
in which one group sees fragments with author names,
while the other only sees the fragments.
After this, a larger experiment should be conducted, 
with more novels, more fragments per novel, and more participants.
It would be interesting to contrast general readers with
readers with particular backgrounds (e.g., literature professors).
Once promising linguistic markers are identified, we want to manipulate
these markers in the fragments, to confirm their effect on literariness,
similar to the work of \citet{hakemulder2004foregrounding}
on novel fragments and \citet{blohm2018sentence} on poetry. 

\section*{Acknowledgments}
We are grateful to Ana Guerberof, Antonio Toral,
and the anonymous reviewers for helpful comments.

\bibliographystyle{aclnatbib}
\bibliography{latech2020}

\end{document}